\documentclass[]{icip}

\usepackage[utf8]{inputenc}
\usepackage[T1]{fontenc}
\usepackage{amsmath}
\usepackage{amsfonts}
\usepackage{amssymb}
\usepackage{booktabs}
\usepackage{caption}
\usepackage{enumitem}
\usepackage{graphicx}
\usepackage{lipsum}
\usepackage{microtype}
\usepackage{multirow}
\usepackage{nicefrac}
\usepackage{subcaption}
\usepackage{tabularx}
\usepackage{url}

\title{From Sequence to Structure: Relational Uncertainty Propagation for LLM Agents}

\author[1,2]{Zhengzhao Ma}
\author[1]{Boxi Cao}
\author[1]{Yaojie Lu}
\author[1]{Hongyu Lin}
\author[1]{Xianpei Han}
\author[1]{Le Sun}

\affiliation[1]{Chinese Information Processing Laboratory, Institute of Software, Chinese Academy of Sciences}
\affiliation[2]{University of Chinese Academy of Sciences, Beijing, China}

\abstract{Reliable uncertainty quantification (UQ) is essential for deploying large language model (LLM) agents in complex interactive environments. Existing UQ methods largely rely on local signals, such as token probabilities, predictive entropy, or per-step confidence, and therefore overlook the long-range dependencies through which errors accumulate across an execution trajectory. As a result, they may fail to identify agent failures whose causes originate several reasoning or interaction steps before the final answer.
We propose RUPA (Relational Uncertainty Propagation for Agents), a trajectory-level UQ framework for LLM agents. RUPA represents an execution history as a directed trajectory graph in which reasoning states, tool interactions, and environment feedback are nodes connected by temporal and semantic dependency edges. It then propagates uncertainty over this graph to capture how execution risk accumulates and transfers across interaction steps. The propagated signal is combined with trajectory-level behavioral features and goal-alignment information to produce a confidence estimate for the full agent trajectory.
We evaluate RUPA on representative agent benchmarks, including $\tau$-2, Terminal-Bench-2, and GAIA, using 6 open-source LLMs spanning multiple model families. Experimental results show that RUPA consistently outperforms existing UQ methods by providing more accurate uncertainty estimates, enabling earlier failure detection, and improving uncertainty-guided agent execution across diverse agent tasks. These results demonstrate that explicitly modeling relational dependency is crucial to reliable UQ for long-horizon LLM agents, providing a practical foundation for trustworthy agent execution.}

\emailresource{{mazhengzhao2024,caoboxi}@iscas.ac.cn}
\code{https://github.com/icip-cas/RUPA}
\begin{document}

\maketitle

\section{Introduction}
\label{sec:introduction}

Large Language Models (LLMs) are rapidly evolving from passive generators into autonomous agents capable of pursuing complex objectives through multi-step reasoning, tool use, and interaction with external environments~\citep{yao2022react,schick2023toolformer,liu2024agentbench}. Such LLM agents have demonstrated remarkable capabilities in software engineering~\citep{jimenez2024swe}, web automation~\citep{zhou2024webarena}, scientific discovery~\citep{zhou2023survey}, and complex decision-making tasks~\citep{mialon2024gaia}. As these systems increasingly perform long-horizon tasks involving tens or even hundreds of reasoning, action, and interaction steps, their reliability has become a critical barrier to real-world deployment~\citep{han2024towards, chen2025consistentchat}. Unlike traditional text generation, failures in LLM agents rarely originate from a single erroneous prediction. Instead, they emerge from the accumulation and propagation of errors across interdependent reasoning steps, tool executions, and environment interactions. Consequently, accurately estimating execution risk before failures occur has become a fundamental challenge for building reliable autonomous agents~\citep{gawlikowski2023survey,yin2024reasoning}.

Uncertainty quantification (UQ) provides a natural solution to this challenge by estimating the reliability of model behavior and enabling downstream strategies such as risk detection, adaptive resampling, self-correction, and decision optimization~\citep{jiang2021can,kadavath2022language,lin2022teaching}. 
However, existing UQ methods are primarily designed for isolated predictions or short-context generation and therefore struggle in long-horizon agent execution. Traditional approaches estimate uncertainty solely from the current model output while largely ignoring risks accumulated throughout the execution history~\citep{manakul2023selfcheckgpt,farquhar2024detecting, mao2026recurrent}. 
More recent agent-oriented UQ methods~\citep{han2024towards,zhao2025uncertainty,kirchhof2025position} begin to incorporate historical information, but they typically model agent trajectories as linear sequences and aggregate uncertainty according to temporal distance or semantic similarity. 
In reality, dependencies between agent steps are inherently relational rather than purely sequential~\citep{jelodar2026integrating, zhang2026confidence, sun2026agentgl}. An early mistake may have little immediate impact unless it influences subsequent reasoning or tool use, whereas a seemingly small misunderstanding can gradually evolve into catastrophic failure if it continuously affects later decisions. 
Without explicitly modeling these dependency structures, uncertainty estimators cannot accurately characterize how execution risks evolve throughout an agent trajectory.

In this work, we argue that uncertainty in LLM agents should be viewed as a trajectory-level property that evolves over the relational structure of agent execution, rather than as a sequence of independent confidence estimates. The key challenge is therefore not simply estimating the uncertainty of each individual step, but understanding how uncertainty propagates through dependencies among reasoning states, actions, tool invocations, and environment observations.

\begin{figure*}[t]
    \centering
    \includegraphics[width=0.9\textwidth]{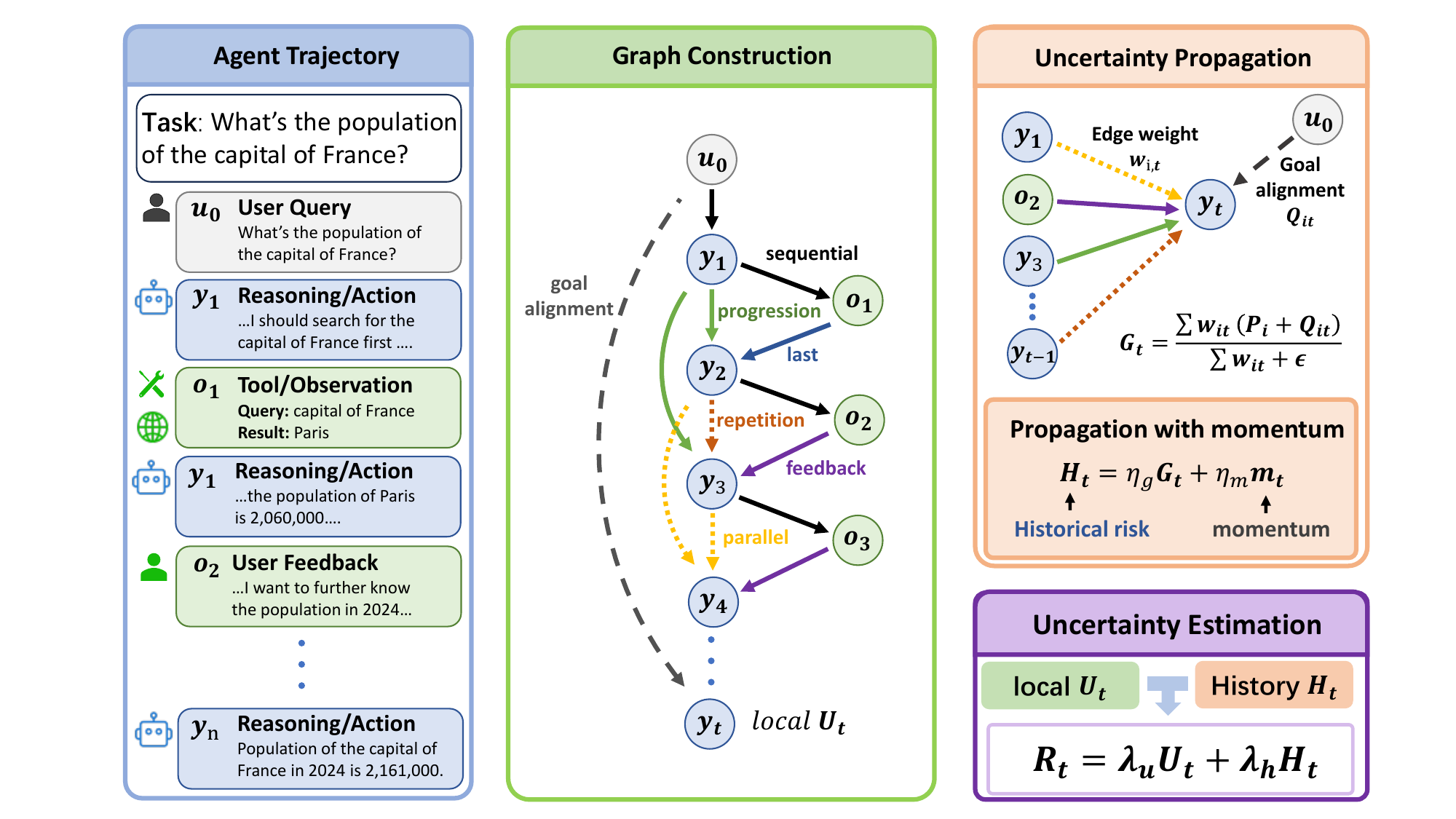}
    \caption{Overview of RUPA. The agent trajectory is converted into a directed dependency graph, uncertainty is propagated over historical states, and the propagated risk is combined with local uncertainty to estimate trajectory-level uncertainty.}
    \label{fig:method}
\end{figure*}

Motivated by this observation, we propose \textbf{R}elational \textbf{U}ncertainty \textbf{P}ropagation for \textbf{A}gents (RUPA), a trajectory-level uncertainty quantification framework for autonomous LLM agents. As shown in Fig~\ref{fig:method}, rather than representing agent execution as a linear sequence, RUPA automatically models the execution trajectory as a directed relational graph, in which nodes represent different execution events, including reasoning states, tool invocations, user interactions, and environment observations. Edges further capture the dependency relations among these nodes, such as sequential transitions, repeated behaviors, reasoning continuation, feedback dependencies, and goal alignment. Based on this structured representation, the uncertainty of each node is jointly determined by its local uncertainty and the historical uncertainty propagated through the directed graph. Specifically, relation-aware edge weights are automatically determined according to the statistical importance of different dependency types, allowing uncertainty to propagate preferentially along influential execution paths. The propagated historical uncertainty is then integrated with the node's local uncertainty to estimate its execution risk. Consequently, RUPA captures the realistic evolution of uncertainty during agent execution, where early critical mistakes continue to influence subsequent dependent reasoning and actions, while uncertainty originating from unrelated execution branches is naturally suppressed. 

We evaluate RUPA on 3 representative agent benchmarks covering diverse LLM agent scenarios, including $\tau$-2~\citep{barres2025tau}, Terminal-Bench-2~\citep{merrill2026terminal}, and GAIA~\citep{mialon2024gaia}, using 6 open-source LLMs ranging from 26B to 230B parameters. Extensive experiments across multiple evaluation settings consistently demonstrate that, compared with existing UQ methods, RUPA substantially improves uncertainty quantification quality, enabling more effective intervention for high-risk agent execution and stronger downstream task performance. First, RUPA achieves the best uncertainty estimation quality on all benchmarks and model families, improving the average AUROC of MiniMax-M2.7 from 0.694 to 0.718 over the strongest baseline. Second, in prefix-based evaluation, RUPA identifies potential execution failures significantly earlier than existing methods, demonstrating a superior early-risk detection capability. Third, RUPA consistently improves downstream task success rates by selecting lower-risk candidate actions during multi-sample decoding. Extensive ablation  studies further show that these improvements primarily originate from relation-aware trajectory graph modeling and uncertainty propagation, highlighting the importance of modeling structural dependencies for reliable uncertainty estimation in autonomous LLM agents.

The main contributions of this work are summarized as\footnote{The code is available at \url{https://github.com/icip-cas/RUPA}.}:
\begin{itemize}
\item We identify relational dependencies between execution steps as a key source of uncertainty evolution in LLM agents, revealing the limitations of existing uncertainty quantification methods that model agent trajectories as independent predictions or linear sequences.
\item We propose \textbf{RUPA}, a graph-based uncertainty quantification framework that represents agent execution as a directed relational graph and performs relation-aware uncertainty propagation to estimate execution risk.
\item Extensive experiments on 3 representative agent benchmarks and 6 LLMs demonstrate that RUPA consistently improves uncertainty estimation quality, early failure detection, and uncertainty-guided agent execution over existing uncertainty quantification methods.
\end{itemize}

\section{Related Works}
\label{sec:related_work}

\subsection{Uncertainty Quantification for LLM}
\label{sec:uq}

Uncertainty quantification (UQ) has become a fundamental component of trustworthy LLMs, aiming to estimate the reliability of model predictions and support downstream tasks~\citep{jiang2021can,kadavath2022language,yan2026denoiseflow}. 
Existing UQ methods for LLMs can be categorized into probability-based, verbalized and sampling-based methods~\citep{yin2024reasoning, heo2024llms}. 

Probability-based approaches~\citep{kossen2024semantic} estimate uncertainty directly from model outputs like predictive entropy, sequence generation probability, or related confidence scores derived from the model's output distribution~\citep{moskvoretskii2025adaptive, li2025confidence}. Verbalized methods prompt models to explicitly output a confidence score alongside the answer~\citep{lin2022teaching, xiong2023can, yang2024verbalized}, offering a flexible and human-interpretable interface~\citep{yoon2025reasoning}. Sampling-based methods~\citep{manakul2023selfcheckgpt, farquhar2024detecting} estimate uncertainty by generating multiple candidate responses and measuring their consistency. They evaluate the agreement among sampled generations to detect hallucinations and factual inconsistencies. 

These methods have demonstrated strong performance~\citep{ding2024retrieve, damani2025beyond,ma2026decoupling}, but they are primarily designed for single-turn prediction. Consequently, they cannot effectively characterize uncertainty across long-horizon reasoning and interaction trajectories~\citep{oh2026uncertainty}.

\subsection{Uncertainty Quantification for LLM Agents} \label{sec:uncertainty_aware_agent}

Agent uncertainty quantification aims to estimate the probability that an entire execution trajectory will successfully accomplish the target task~\citep{kirchhof2025position,zhang2026selaur}. Several recent methods extend traditional uncertainty estimation from individual responses to complete agent trajectories, including SAUP~\citep{zhao2025uncertainty}, Tracer~\citep{tayebati2026tracer}, and UProp~\citep{duan2025uprop}. These methods improve uncertainty estimation compared with conventional approaches and demonstrate the importance of utilizing execution history~\citep{shi2026confidence}. 

However, existing agent UQ methods predominantly represent execution trajectories as linear sequences. As a result, the underlying dependency structure among reasoning steps remains largely unexplored. Ignoring these relational dependencies makes it difficult to accurately capture how execution risks accumulate and propagate throughout long-horizon agent trajectories~\citep{li2026trajectories}. 

\section{Empirical Analysis of Agent Uncertainty}
\label{sec:empirical}

\begin{table}[t]
\centering
\begin{tabular}{lccc}
\toprule
\textbf{Domain} & \textbf{Random} & \textbf{Seq. Prob.} & \textbf{Verbalize} \\
\midrule
Airline  & 0.441 & 0.205 & 0.485 \\
Retail   & 0.472 & 0.301 & 0.523 \\
\bottomrule
\end{tabular}
\caption{Preliminary AUROC evaluation of traditional UQ methods on $\tau^2$ agent tasks.}
\label{tab:preliminary_uq_tau2}
\end{table}

\begin{figure}[t]
\begin{center}
\includegraphics[width=0.28\textwidth]{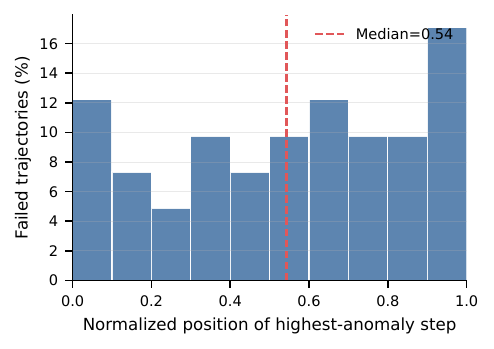}
\includegraphics[width=0.35\textwidth]{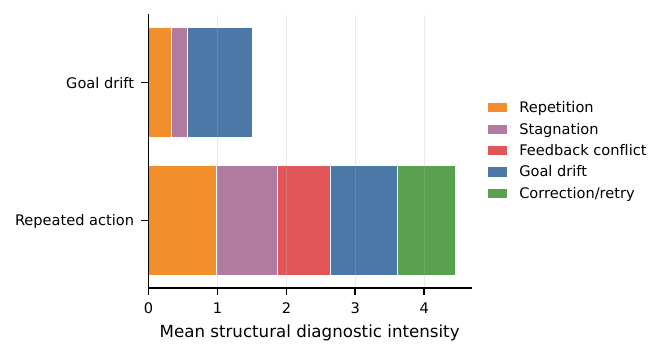}
\end{center}
\caption{Structural analysis of failure-indicative signals in failed agent trajectories.}
\label{fig:failure}
\end{figure}

We first investigate why uncertainty estimation for agents is fundamentally more challenging than conventional LLM tasks. Through empirical analysis, we show that execution failures are often induced by relational dependencies distributed across the entire trajectory.
These observations motivate trajectory-level relational uncertainty modeling for reliable agent uncertainty quantification.

\paragraph{Traditional uncertainty estimation fails on long-horizon agent tasks.}
\label{sec:traditional_uq_struggles}
We first evaluate whether UQ methods designed for conventional language generation remain effective in long-horizon agent tasks.
Specifically, we conduct a preliminary study on representative $\tau$-2 domains, \textit{Airline} and \textit{Retail}, using Qwen3.5-27B model.
We compare two widely adopted UQ methods, sequence probability and verbalized confidence, for trajectory-level failure prediction.

As shown in Table~\ref{tab:preliminary_uq_tau2}, both methods perform poorly, with AUROC values close to random guessing across the two domains.
For example, on the \textit{Airline} domain, sequence probability obtains an AUROC of only 0.205, while verbalized confidence reaches 0.485.
Similar observations hold for the \textit{Retail} domain.
These results indicate that uncertainty estimated solely from local generation confidence is insufficient for identifying failures in long-horizon agent execution. This suggests that execution failures depend on information beyond the current generation, motivating a closer examination of where failure-indicative uncertainty originates.

\paragraph{Failure signals are distributed over relational trajectory dependencies.}
To better understand why traditional uncertainty estimation fails, we analyze where failure-indicative signals emerge within agent trajectories.
For each failed trajectory, we compute a step-level risk score and identify the execution step with the highest estimated anomaly. Then we analyze its relative position and dependency relation profile.

As shown in Fig~\ref{fig:failure}, high-risk steps are distributed throughout the execution trajectory rather than concentrating near the final answer, which suggests that failures often originate from intermediate reasoning or interaction steps and gradually propagate to subsequent decisions. Furthermore, the failure step exhibits an average repetition score of 0.981 and a stagnation score of 0.883, while feedback-conflict and correction/retry relations also appear frequently.
These patterns indicate that execution failures are associated with relational, structural dependencies among trajectories.

Overall, these observations reveal that uncertainty in agent execution is inherently trajectory-dependent.
Consequently, modeling execution trajectories as linear sequences is insufficient to accurately characterize risk evolution.
This empirical evidence motivates the graph-based uncertainty propagation framework proposed in the following section.

\section{Methods}
\label{sec:methods}

Motivated by the above insight, we propose \textbf{R}elational \textbf{U}ncertainty \textbf{P}ropagation for \textbf{A}gents (RUPA), a relational trajectory-aware UQ framework for LLM agents.
Fig~\ref{fig:method} presents an overview of RUPA. Given an agent execution trajectory, RUPA converts the execution trajectory into a relational dependency graph. Then it propagates uncertainty through the edges of the graph to capture how execution risks accumulate across reasoning, tool use, and environment interaction. Finally, the propagated structural uncertainty is combined with the local uncertainty of the current reasoning step to produce a trajectory-aware uncertainty estimate. 

\subsection{Relational Trajectory Graph Construction}
\label{sec:graph_construction}

To capture the relational trajectory-dependencies in agent uncertainty quantification, RUPA represents each execution prefix as a directed trajectory graph $\mathcal{G} = (\mathcal{V}, \mathcal{E})$. Each node $i \in \mathcal{V}$ corresponds to an execution event, including user instructions, assistant reasoning or actions, tool invocations, and environment observations. Directed edges $e_{i,j} \in \mathcal{E}$ describe logical dependencies between historical events and the current reasoning step. RUPA constructs only dependency-related edges within a bounded historical context. We consider seven representative relation types:

\begin{itemize} 
\item \textbf{Sequential}: the immediately preceding execution.
\item \textbf{Latest}: the most recent environment or user instructions. 
\item \textbf{Repetition}: actions exhibiting highly similar reasoning patterns or tool usage.
\item \textbf{Progression}: reasoning steps extending an existing solution step.
\item \textbf{Parallel}: alternative reasoning branches under the same task context.
\item \textbf{Feedback}: feedback to environment observations indicating execution failures or unstable outputs.
\item \textbf{Goal Alignment}: semantic dependency between the current reasoning step and the original task objective. 
\end{itemize}
The type of the edges is determined by features like embedding distance and matching cues. 
Together, these relations characterize how execution states interact throughout an agent trajectory, providing a structured representation for uncertainty propagation beyond simple temporal ordering. 
The detailed edge construction is shown in the Appendix.

\subsection{Relation-aware Uncertainty Propagation} 
\label{sec:graph_propagation}

Given the trajectory graph $\mathcal{G}$, RUPA propagates uncertainty from historical execution states to the current node. Unlike conventional sequential aggregation, RUPA uncertainty propagation is guided by dependency relations in the graph.

In particular, each node holds a local uncertainty $U_{t}$. For assistant nodes, local uncertainty is computed from predictive entropy. For environment nodes, uncertainty is estimated from observable interaction signals, including execution failures, empty tool responses, and conflicting environment feedback. These uncertainty values serve as the initial risk associated with each graph node. 

Since different dependency relations should contribute unequally to possible future failures, RUPA learns a propagation weight for each graph edge, where relation types exhibiting stronger structural variation across trajectories receive larger propagation coefficients. The propagation weight of each edge is then determined by its relation reliability, relation strength, and temporal distance:

\begin{equation}
w_{it}
=
\rho_{\tau_{it}}
\,
\tilde r_{it}
\,
\delta^{\mathrm{age}(i,t)-1},
\end{equation}

where $\delta$ is the temporal decay factor. Consequently, structurally important historical dependencies exert greater influence, while obsolete execution states are gradually discounted. Specially, for goal-alignment edge, instead of computing an edge weight, we estimate a goal alignment score:
\begin{equation}
    Q_{it} = 1 - S(y_t, x)
\end{equation}
where $S(y_{t}, x)$ is a similarity function.

Finally, RUPA propagates uncertainty over the trajectory graph to estimate the structural risk of the current reasoning step. The propagated uncertainty is computed by aggregating uncertainty from all dependency-related historical nodes,

\begin{equation}
G_t=
\frac{
\sum_{i\in\mathcal N(t)}
w_{it}
(P_i+Q_{it})
}{
\sum_{i\in\mathcal N(t)}
w_{it}
+\epsilon},
\end{equation}

where $P_i$ denotes the propagated uncertainty stored at historical node $i$, and $\mathcal N(t)$ denotes the neighboring historical nodes connected to the current step. Besides graph propagation, RUPA also maintains an exponentially decayed uncertainty momentum to preserve long-range execution trends,

\begin{equation}
m_t=
\frac{
\sum_{k<t}\gamma^{t-k}P_k
}{
\sum_{k<t}\gamma^{t-k}
+\epsilon},
\qquad
H_t
=
\eta_g G_t
+
\eta_m m_t,
\end{equation}

where $H_t$ denotes the propagated historical uncertainty. This formulation allows uncertainty to accumulate along dependency paths throughout the trajectory, enabling early execution failures to continuously influence subsequent reasoning even when the current generation itself appears confident.

RUPA combines the intrinsic uncertainty of the current generation $U_{t}$ and the structural uncertainty accumulated throughout the history $H_{t}$ by a simple additive formulation: 
\begin{equation} 
R_t = \lambda_u U_t + \lambda_h H_t, 
\end{equation} 

The resulting score $R_t$ serves as the uncertainty estimate of the current reasoning step and is propagated to subsequent execution states. 
For complete trajectories, RUPA aggregates the step-level uncertainty scores to obtain a trajectory-level uncertainty estimate, where larger scores indicate a higher probability of task failure.

\section{Experiments}
\label{sec:experiments}

In this section, we evaluate RUPA on multi-turn agent tasks. The experiments assess both uncertainty estimation quality and the effect of uncertainty estimates on downstream agent execution.

\subsection{Experimental Settings}
\label{sec:experimental_settings}

\textbf{Datasets.}
We evaluate all methods on 3 representative agent benchmarks: $\tau$-2, Terminal-Bench-2, and GAIA, covering conversational decision making, terminal-based software engineering, and open-domain complex problem solving.
Together, they provide a comprehensive evaluation of UQ methods under different reasoning and interaction patterns.

\textbf{Baselines.}
We compare RUPA against five representative uncertainty quantification methods.
\textbf{PE} estimates uncertainty using predictive entropy aggregated over the trajectory.
\textbf{SP} uses sequence generation probability as a confidence signal.
\textbf{SAUP} estimates uncertainty from scene-aware execution risks.
\textbf{Tracer} performs trajectory-level UQ by aggregating interaction signals across the execution process.
\textbf{UProp} models uncertainty propagation using pointwise mutual information.
Together, these baselines cover both conventional token-level uncertainty estimation and recent trajectory-aware agent uncertainty quantification methods. The quality of each UQ method is evaluated with AUROC, AUPRC, and the best F1 score over all decision thresholds. 

\textbf{Models.}
Experiments are conducted using 6 representative open-source LLMs spanning multiple model families and scales:
Qwen3.5-27B,
Qwen3.6-35B-A3B,
Gemma-4-26B-it,
Gemma-4-31B-it,
GPT-OSS-120B,
and MiniMax-M2.7.
These models provide complete reasoning traces and token-level probabilities required by all UQ methods.

\begin{table*}[t]
\centering
\small
\setlength{\tabcolsep}{3 pt}
\resizebox{\textwidth}{!}{
\begin{tabular}{clcccccccccccc}
\toprule
\multirow{2}{*}{\textbf{Model}} &
\multirow{2}{*}{\textbf{Method}}
&
\multicolumn{3}{c}{\textbf{Average}}
&
\multicolumn{3}{c}{\textbf{$\tau$-2}}
&
\multicolumn{3}{c}{\textbf{Terminal-Bench}}
&
\multicolumn{3}{c}{\textbf{GAIA}} \\
\cmidrule(lr){3-5}
\cmidrule(lr){6-8}
\cmidrule(lr){9-11}
\cmidrule(lr){12-14}
&
&
AUROC & AUPRC & F1
&
AUROC & AUPRC & F1
&
AUROC & AUPRC & F1
&
AUROC & AUPRC & F1
\\
\midrule
\multirow{6}{*}{Qwen3.5-27B}
& Entropy
& 0.559 & 0.543 & 0.538
& 0.573 & 0.623 & 0.438
& 0.508 & 0.679 & 0.705
& 0.595 & 0.326 & 0.470 \\
& Seq-prob
& 0.566 & 0.612 & 0.582
& 0.547 & 0.636 & 0.509
& 0.541 & 0.682 & 0.667
& 0.611 & 0.517 & 0.571 \\
& SAUP
& 0.595 & 0.561 & 0.630
& 0.630 & 0.681 & 0.595
& 0.491 & 0.611 & \textbf{0.763}
& 0.665 & 0.390 & 0.533 \\
& Tracer
& 0.608 & 0.559 & 0.641
& 0.634 & 0.678 & 0.649
& 0.511 & 0.605 & 0.730
& 0.680 & 0.394 & 0.545 \\
& Uprop
& 0.588 & 0.623 & 0.645
& 0.628 & 0.691 & 0.677
& 0.530 & 0.668 & 0.671
& 0.606 & 0.511 & 0.587 \\
& \textbf{RUPA}
& \textbf{0.656} & \textbf{0.746} & \textbf{0.687}
& \textbf{0.677} & \textbf{0.733} & \textbf{0.714}
& \textbf{0.594} & \textbf{0.725} & 0.736
& \textbf{0.697} & \textbf{0.781} & \textbf{0.611} \\
\midrule
\multirow{6}{*}{Qwen3.6-35B}
& Entropy
& 0.579 & 0.627 & 0.583
& 0.631 & 0.717 & 0.706
& 0.569 & 0.671 & 0.622
& 0.538 & 0.493 & 0.421 \\
& Seq-prob
& 0.596 & 0.583 & 0.616
& 0.644 & 0.709 & 0.694
& 0.610 & 0.672 & 0.693
& 0.534 & 0.367 & 0.462 \\
& SAUP
& 0.608 & 0.644 & 0.673
& 0.651 & 0.733 & 0.735
& 0.628 & 0.713 & 0.790
& 0.545 & 0.487 & 0.493 \\
& Tracer
& 0.629 & 0.663 & 0.677
& 0.647 & 0.727 & 0.736
& 0.649 & 0.695 & 0.794
& 0.590 & 0.566 & 0.500 \\
& Uprop
& 0.595 & 0.595 & 0.661
& 0.612 & 0.711 & 0.720
& 0.651 & 0.714 & 0.795 
& 0.523 & 0.360 & 0.467 \\
& \textbf{RUPA}
& \textbf{0.645} & \textbf{0.679} & \textbf{0.694}
& \textbf{0.670} & \textbf{0.741} & \textbf{0.759}
& \textbf{0.657} & \textbf{0.724} & \textbf{0.805}
& \textbf{0.608} & \textbf{0.571} & \textbf{0.518} \\
\midrule
\multirow{6}{*}{Gemma4-26B}
& Entropy
&0.711&0.779&0.716
&0.712&0.784&0.776
&0.643&0.701&0.646
&0.779&0.851&0.726 \\
& Seq-prob
& 0.712 & 0.838 & 0.759
& 0.701 & 0.810 & 0.761
& 0.580 & 0.769 & 0.683
& 0.854 & 0.934 & 0.833 \\
& SAUP
&0.761&0.820&0.775
&0.731&0.815&0.814
&0.677&0.715&0.671
&\textbf{0.874}&0.930&0.839 \\
& Tracer
&0.743&0.819&0.732
&0.720&0.797&0.800
&0.669&0.730&0.668
&0.841&0.929&0.729 \\
& Uprop
& 0.721 & 0.835 & 0.758
& 0.718 & 0.804 & 0.792
& 0.585 & 0.768 & 0.649
& 0.861 & 0.932 & 0.833 \\
& \textbf{RUPA}
&\textbf{0.780}&\textbf{0.851}&\textbf{0.784}
&\textbf{0.756}&\textbf{0.823}&\textbf{0.819}
&\textbf{0.713}&\textbf{0.781}&\textbf{0.690}
& 0.872&\textbf{0.949}&\textbf{0.842} \\
\midrule
\multirow{6}{*}{Gemma4-31B}
& Entropy
&0.806&0.919&0.782
&0.815&0.887&0.772
&0.728&0.916&0.714
&0.875&0.954&0.861 \\
& Seq-prob
& 0.818 & 0.910 & 0.806
& 0.827 & 0.870 & 0.764
& 0.740 & 0.915 & 0.783
& 0.886 & 0.944 & 0.871 \\
& SAUP
&0.842&0.935&0.844
&0.833&0.905&0.828
&0.802&0.936&0.826
&0.890&0.963&0.879 \\
& Tracer
&0.823&0.920&0.777
&0.763&0.858&0.741
&0.807&0.937&0.735
&\textbf{0.900}&0.965&0.855 \\
& Uprop
& 0.836 & 0.918 & 0.842
& 0.887 & 0.897 & 0.841
& 0.751 & 0.918 & 0.832
& 0.869 & 0.940 & 0.854 \\
& \textbf{RUPA}
&\textbf{0.861}&\textbf{0.949}&\textbf{0.882}
&\textbf{0.877}&\textbf{0.936}&\textbf{0.914}
&\textbf{0.815}&\textbf{0.941}&\textbf{0.838}
&0.892&\textbf{0.971}&\textbf{0.895} \\
\midrule
\multirow{6}{*}{GPT-oss-120B}
& Entropy
&0.492&0.819&0.726
&0.491&0.751&0.646
&0.495&0.904&0.737
&0.489&0.804&\textbf{0.794} \\
& Seq-prob
& 0.464 & 0.843 & 0.427
& 0.488 & 0.792 & 0.417
& 0.299 & 0.867 & 0.121
& 0.605 & 0.870 & 0.742 \\
& SAUP
&0.520&0.870&0.765
&0.512&0.869&0.732
&0.486&0.902&0.801
&0.563&0.840&0.761 \\
& Tracer
&0.525&0.875&0.393
&0.528&0.860&0.457
&0.419&0.898&0.228
&0.629&0.867&0.494 \\
& Uprop
& 0.567 & 0.858 & 0.504
& 0.716 & 0.831 & 0.689
& 0.343 & 0.871 & 0.211
& 0.642 & 0.872 & 0.612 \\
& \textbf{RUPA}
&\textbf{0.577}&\textbf{0.889}&\textbf{0.815}
&\textbf{0.577}&\textbf{0.884}&\textbf{0.762}
&\textbf{0.500}&\textbf{0.907}&\textbf{0.902}
&\textbf{0.655}&\textbf{0.875}&0.781 \\
\midrule
\multirow{6}{*}{MiniMax-M2.7}
& Entropy
&0.666&0.749&0.689
&0.651&0.788&0.675
&0.635&0.800&0.709
&0.712&0.659&0.682 \\
& Seq-prob
& 0.685 & 0.748 & 0.718
& 0.683 & 0.815 & 0.697
& 0.669 & 0.799 & 0.737
& 0.704 & 0.630 & 0.721 \\
& SAUP
&0.670&0.772&0.701
&0.626&0.817&0.706
&0.671&0.826&0.720
&0.714&0.674&0.675 \\
& Tracer
&0.694&0.778&0.745
&0.677&0.832&0.735
&0.679&0.825&0.771
&0.725&0.675&0.729 \\
& Uprop
& 0.669 & 0.750 & 0.710
& 0.625 & 0.821 & 0.726
& 0.672 & 0.798 & 0.685
& 0.711 & 0.632 & 0.720 \\
& \textbf{RUPA}
&\textbf{0.718}&\textbf{0.805}&\textbf{0.755}
&\textbf{0.728}&\textbf{0.840}&\textbf{0.738}
&\textbf{0.690}&\textbf{0.849}&\textbf{0.788}
&\textbf{0.736}&\textbf{0.727}&\textbf{0.739} \\
\bottomrule
\end{tabular}
}
\caption{Uncertainty quantification performance across agent benchmarks. The best result for each model is shown in \textbf{bold}.}
\label{tab:main}
\end{table*}

\subsection{Overall Performance}
\label{sec:main_result}

Table~\ref{tab:main} summarizes uncertainty quantification performance across agent tasks and model families.
Overall, our proposed RUPA consistently achieves the best uncertainty estimation performance across different model families, demonstrating that explicitly modeling trajectory-level dependency and uncertainty propagation is an effective solution for failure detection in long-horizon agent execution.

\paragraph{Traditional UQ methods struggle in long-horizon, interactive agent environments.}
Traditional uncertainty quantification methods, including entropy and sequence probability, consistently underperform across nearly all settings.
In particular, on Qwen3.5-27B, Entropy achieves only 0.559 AUROC on average, compared with 0.656 achieved by RUPA. A similar trend can be observed on GPT-OSS-120B, where Entropy obtains only 0.492 AUROC while RUPA improves it to 0.577.
These results demonstrate that traditional UQ methods only exploit the confidence of the current generation and ignore dependencies introduced by previous reasoning and environment interactions, making them inadequate for multi-turn agent trajectories.

\paragraph{Sequential agent UQ methods can not fully detect risk propagation in complex step relations.}
As shown in Table~\ref{tab:main}, recent agent-oriented uncertainty quantification methods improve over traditional confidence-based approaches by incorporating sequential execution information while they fail to capture relational structure based risk propagation.
For instance, on Qwen3.6-35B, Tracer achieves 0.629 average AUROC, while RUPA further improves it to 0.645. Similarly, on Gemma4-26B, the SAUP baseline reaches 0.761 AUROC, whereas RUPA increases this score to 0.780.
The improvement becomes even more evident on $\tau$-2 and Terminal-Bench, where modeling relation-aware uncertainty propagation enables more accurate identification of failures accumulated over long execution trajectories.

\paragraph{RUPA achieves a favorable performance in agent trajectory failure detection.} Across all six evaluated models, RUPA achieves the highest average AUROC, AUPRC, and F1 score while consistently outperforming previous agent UQ approaches on individual benchmarks.
In particular, RUPA improves the average AUROC from 0.608 to 0.656 on Qwen3.5-27B, from 0.629 to 0.645 on Qwen3.6-35B, from 0.761 to 0.780 on Gemma4-26B, from 0.842 to 0.861 on Gemma4-31B, and from 0.694 to 0.718 on MiniMax-M2.7.
These results demonstrate the effectiveness and strong generalization ability of trajectory graph modeling and relational uncertainty propagation for agent uncertainty estimation.

\subsection{Detailed Analysis}
\label{sec:detailed_analysis}

\begin{figure*}[t]
\begin{center}
\includegraphics[width=0.24\textwidth]{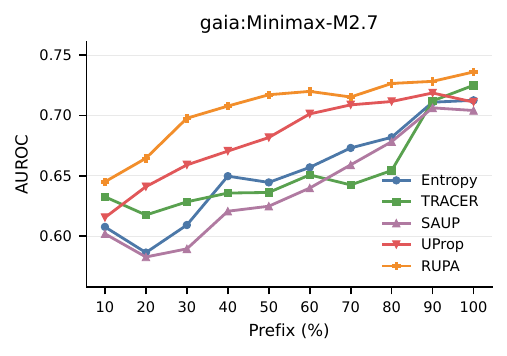}
\includegraphics[width=0.24\textwidth]{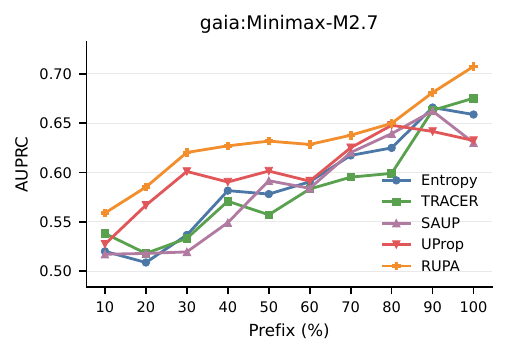}
\includegraphics[width=0.24\textwidth]{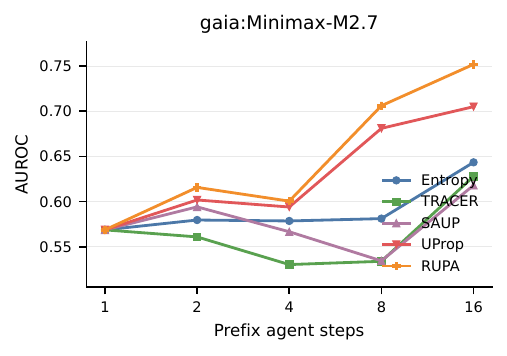}
\includegraphics[width=0.24\textwidth]{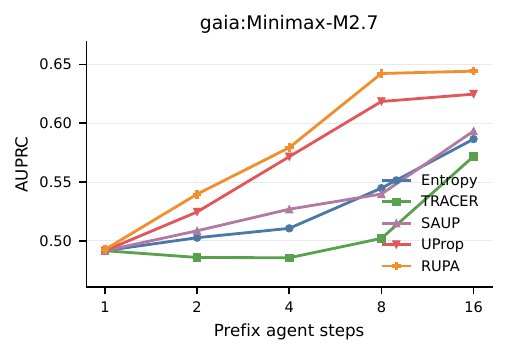}
\includegraphics[width=0.24\textwidth]{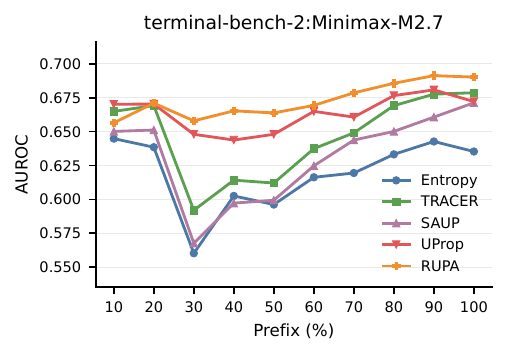}
\includegraphics[width=0.24\textwidth]{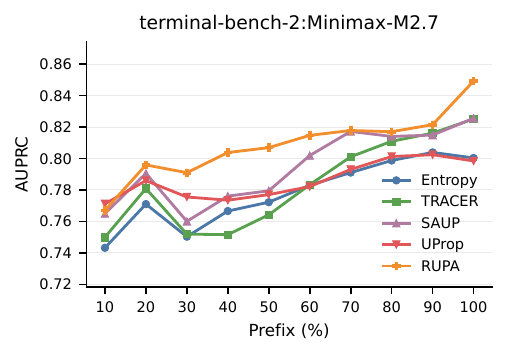}
\includegraphics[width=0.24\textwidth]{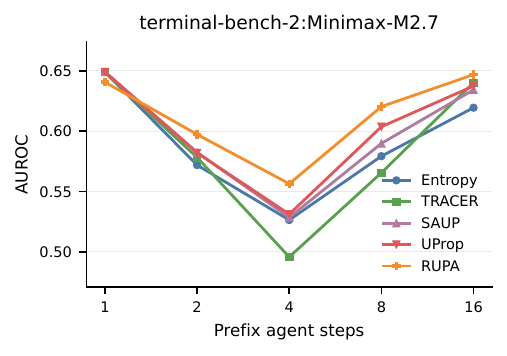}
\includegraphics[width=0.24\textwidth]{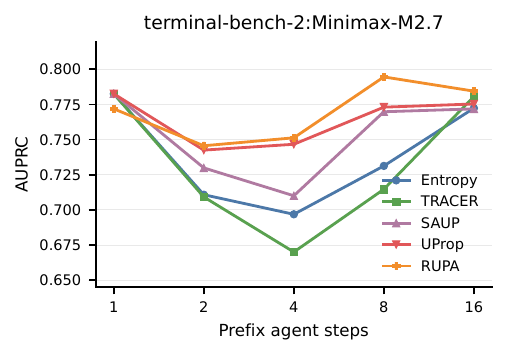}
\end{center}
\caption{Prefix-based early failure detection on GAIA and Terminal-Bench with MiniMax-M2.7. Curves report AUROC and AUPRC when each method observes only a fixed percentage or a fixed number of steps from the trajectory prefix.}
\label{fig:sample}
\end{figure*}

\paragraph{RUPA enables earlier failure detection with partial trajectories.}
A practical UQ method should identify failure risks before the agent completes its entire execution process. To evaluate this capability, we conduct a prefix-based analysis in which each trajectory is truncated by retaining a fixed percentage or number of reasoning/action steps. Uncertainty estimation is then performed using only the partial trajectory available.
Fig~\ref{fig:sample} demonstrates that RUPA consistently achieves higher uncertainty prediction performance. In particular, the advantage is particularly pronounced when only a small or moderate fraction of the trajectory is observed, which suggests that the propagated uncertainty signals emerge early during agent execution and can be leveraged to anticipate future failures before execution finished.

\begin{table}[t]
\centering
\small
\setlength{\tabcolsep}{15pt}
\begin{tabular}{clccc}
\toprule
\textbf{Model} & \textbf{Method} & \textbf{TB2} & \textbf{GAIA} \\
\midrule
\multirow{5}{*}{Qwen3.5-27B}
& Random & 0.105 & 0.261 \\
& Entropy& 0.141 & 0.282 \\
& SAUP   & 0.150 & 0.273 \\
& Tracer & 0.143 & 0.267 \\
& \textbf{Ours}& \textbf{0.213} & \textbf{0.297} \\
\midrule
\multirow{5}{*}{Gemma4-31B}
& Random & 0.122 & 0.175 \\
& Entropy& 0.143 & 0.206 \\
& SAUP   & 0.146 & 0.221 \\
& Tracer & 0.149 & 0.229 \\
& \textbf{Ours}& \textbf{0.213} & \textbf{0.242} \\
\midrule
\multirow{5}{*}{GPT-OSS-120B}
& Random & 0.079 & 0.127 \\
& Entropy& 0.112 & 0.140 \\
& SAUP   & 0.124 & 0.157 \\
& Tracer & 0.167 & 0.188 \\
& \textbf{Ours}& \textbf{0.202} & \textbf{0.242} \\
\midrule
\multirow{5}{*}{MiniMax-M2.7}
& Random & 0.225 & 0.284 \\
& Entropy& 0.240 & 0.299 \\
& SAUP   & 0.247 & 0.285 \\
& Tracer & 0.259 & 0.316 \\
& \textbf{Ours}& \textbf{0.270} & \textbf{0.339} \\
\bottomrule
\end{tabular}
\caption{Uncertainty-guided sampling agent performance.}
\label{tab:downstream}
\end{table}

\begin{figure}[t]
\begin{center}
\includegraphics[width=0.45\textwidth]{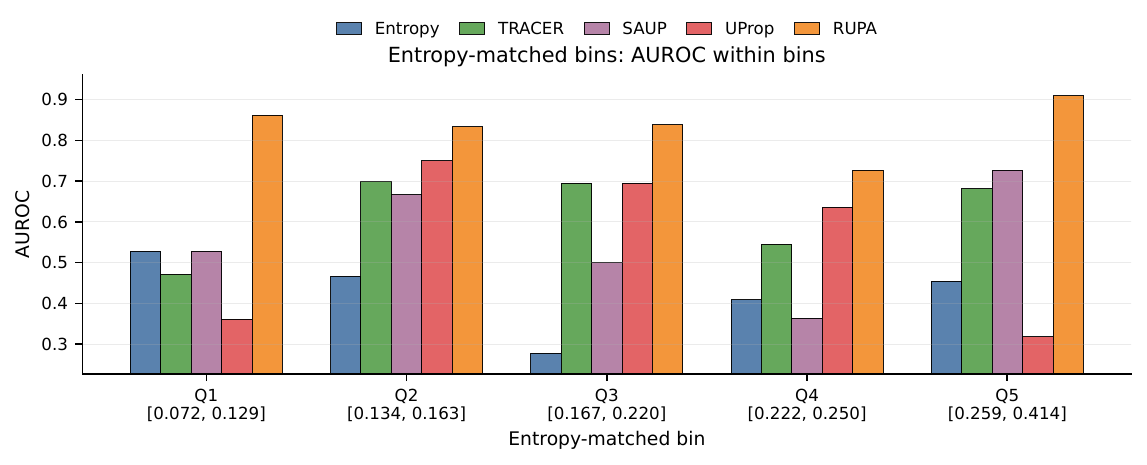}
\includegraphics[width=0.45\textwidth]{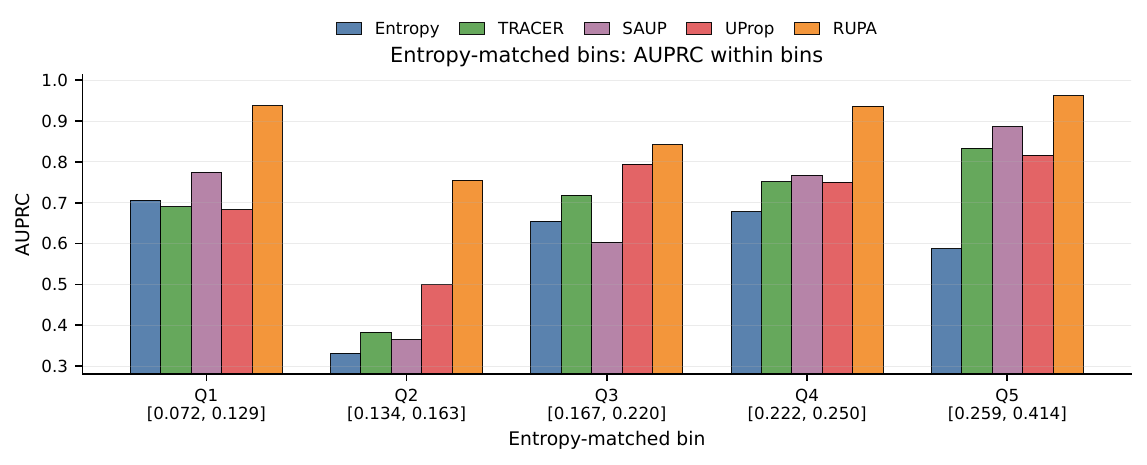}
\end{center}
\caption{Entropy-matched analysis of trajectory-graph confidence signals.}
\label{fig:bins}
\end{figure}

\paragraph{RUPA's uncertainty is able to translate into better agent performance.}
To investigate whether improved uncertainty estimation can benefit agent execution, we built a trivial uncertainty-guided agent framework. At each decision step, the agent samples multiple candidate actions and selects the action associated with the lowest predicted uncertainty score. We evaluate this strategy on Terminal-Bench-2 and GAIA. Table~\ref{tab:downstream} shows the final task accuracy, showing that RUPA consistently achieves the strongest downstream performance across all evaluated models and benchmarks. In particular, on Terminal-Bench-2, the accuracy of Qwen3.5-27B improves from 0.105 under random selection to 0.213 when guided by RUPA. Comparable gains are observed on GAIA, where RUPA consistently outperforms entropy-based and agent-specific uncertainty baselines. These results demonstrate that uncertainty estimates produced by RUPA are not only more accurate for failure detection but also actionable for improving agent decision making.

\paragraph{Graph-based trajectory modeling provides complementary uncertainty signals.}
To understand whether graph-based trajectory modeling of RUPA provides additional uncertainty information than traditional confidence estimation alone, we compare uncertainty prediction performance under entropy-controlled settings.
Specifically, we divide trajectories into bins with similar entropy-based uncertainty scores on GAIA trajectories generated by MiniMax-M2.7 and evaluate whether graph propagation can still distinguish successful and failed executions.
Fig~\ref{fig:bins} shows that graph-based uncertainty remains highly informative even when entropy values are nearly identical.
In low-entropy regions (e.g., Q1), conventional probability-based uncertainty estimators achieve AUROC performance of about 0.5 because local token confidence is highly similar across trajectories. In contrast, RUPA still achieves an AUROC of approximately 0.85 and substantially improves AUPRC to around 0.93.
These results indicate that problematic dependency structures are sufficient in agent uncertainty propagation. Therefore, explicitly modeling trajectory relations provides complementary information beyond token-level confidence signals.

\subsection{Ablation Study}
\label{sec:ablation}

To understand the contribution of each component in RUPA, we conduct ablation studies on the MiniMax-M2.7 model. Table~\ref{tab:ablation} reports the uncertainty estimation performance under different ablation settings.

\begin{table}[t]
\centering
\small
\setlength{\tabcolsep}{4pt}
\begin{tabular}{lccc}
\toprule
\textbf{Method} & \textbf{AUROC} & \textbf{AUPRC} & \textbf{F1} \\
\midrule
Full RUPA & \textbf{0.718} & \textbf{0.805} & \textbf{0.755}\\
w/o graph modeling & 0.678 & 0.642 & 0.675\\
w/o propagation & 0.689 & 0.657 & 0.694\\
w/o 50\% graph edges & 0.705 & 0.654 & 0.739\\
w/ random graph & 0.681 & 0.643 & 0.691\\
\bottomrule
\end{tabular}
\caption{Ablation study of RUPA on 3 agent task benchmarks.}
\label{tab:ablation}
\end{table}

Table~\ref{tab:ablation} shows that both relational graph modeling and uncertainty propagation contribute substantially to the performance of RUPA. In particular, removing graph modeling leads to the largest performance degradation, reducing AUROC from 0.718 to 0.678 and AUPRC from 0.805 to 0.642, which demonstrates that representing agent trajectories as dependency graphs is essential for capturing failure-indicative structural information beyond local uncertainty estimates. When uncertainty propagation is further removed, performance also drops noticeably, demonstrating that explicitly propagating uncertainty across related execution states is critical for modeling long-range failure accumulation.

Furthermore, replacing the relational graph with a random topology results in a similar performance degradation to removing graph modeling, indicating that the performance gains arise from meaningful dependency structures rather than simply introducing additional graph features.

\section{Conclusion}
\label{conclusion}

This paper studies uncertainty quantification for long-horizon LLM agents. We argue that uncertainty in agent execution arises from the dependency structure among reasoning steps, tool interactions, and environment feedback, rather than from isolated model predictions. 

Motivated by this observation, we propose RUPA, a relation-aware uncertainty quantification framework that represents agent trajectories as dependency graphs and propagates uncertainty along meaningful execution relations to capture the accumulation of failure risks. Extensive experiments demonstrate that RUPA consistently outperforms both conventional uncertainty quantification methods and recent agent-specific baselines. RUPA also enables earlier failure detection and consistently improves downstream uncertainty-guided agent performance, highlighting the practical value of trajectory-aware uncertainty estimation for reliable autonomous agents.

\bibliographystyle{plainnat}
\bibliography{paper}
\clearpage
\appendix
\section{Appendix}
\subsection{Implementation Details of RUPA}

RUPA uses lightweight deterministic detectors to construct trajectory edges from observable prefix information. Each trajectory step is first normalized into a textual state representation by concatenating the assistant message, reasoning content, tool-call signature, and observation text when available. The resulting text is tokenized after lowercasing, punctuation removal, stop-word filtering, and numeric-token filtering. Tool calls are canonicalized as function-name--argument signatures, which allows RUPA to compare repeated tool usage across steps.

For a candidate historical node $v_i$ and the current assistant node $v_t$, RUPA computes token and tool-use matching scores by text embedding distances. In our implementation, a bge-m3 model serve as the embedding model. An alternative way is to compute token overlap in case no embedding models available.

In addition to matching scores, RUPA uses lexical cue matching to distinguish logical relations. 
Progression edges are detected by matching continuation or refinement cues such as \textit{next}, \textit{therefore}, \textit{continue}, \textit{verify}, and \textit{test}. Parallel edges are detected by alternative-branch cues such as \textit{alternative}, \textit{instead}, \textit{another}, \textit{different}, \textit{try}, and \textit{fallback}. Feedback edges
are detected when previous observations contain instability cues, including empty observations, \textit{traceback}, \textit{error}, \textit{exception}, \textit{failed} or \textit{timeout}. This design
avoids using future outcomes or final verifier labels during graph construction.

Specifically, for each relation edge type $\tau$, we compute the edge weight by reliablity and its relation strength. The reliability coefficient is computed from unlabeled training trajectories based on the variation of its normalized relation strength,

\begin{equation}
q_\tau=
\frac{\mathrm{Var}(\tilde r_\tau)}
{\mathbb{E}(\tilde r_\tau)+\epsilon},
\qquad
\rho_\tau=
|\mathcal T|
\frac{\exp(q_\tau/T)}
{\sum_{\tau'\in\mathcal T}\exp(q_{\tau'}/T)},
\end{equation}

where $\mathcal T$ denotes the set of relation types and $T$ is a temperature parameter. 

The detailed hyperparameters of RUPA is shown in table ~\ref{tab:rupa_hyperparams} and Table~\ref{tab:rupa_edge_caps}

\begin{table}
\centering
\begin{tabular}{lc}
\toprule
\textbf{Hyperparameter} & \textbf{Default}  \\
\midrule
Visible Window & 5  \\
weight decay $\delta$ & 0.80  \\
momentum decay $\gamma$ & 0.75  \\
Graph propagation weight $\eta_g$ & 0.70 \\
momentum weight $\eta_{\mathrm{m}}$ & 0.30  \\
History weight $\lambda_h$ & 1.0 \\
local weight$\lambda_u$ & 1.0  \\
\bottomrule
\end{tabular}
\caption{Default hyperparameter settings used in RUPA.}
\label{tab:rupa_hyperparams}
\end{table}

\begin{table}
\centering
\begin{tabular}{lc}
\toprule
\textbf{Edge type} & \textbf{relation strength} \\
\midrule
Sequential & 0.35 \\
Latest & 0.75 \\
Repetition & 0.95 \\
Feedback & 0.85 \\
Progression & 0.65 \\
Parallel & 0.45 \\
\bottomrule
\end{tabular}
\caption{Type-specific edge strength used in RUPA.}
\label{tab:rupa_edge_caps}
\end{table} 

\begin{figure}
\begin{center}
\includegraphics[width=0.3\textwidth]{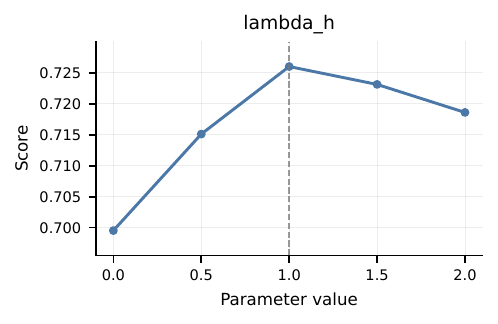}
\includegraphics[width=0.3\textwidth]{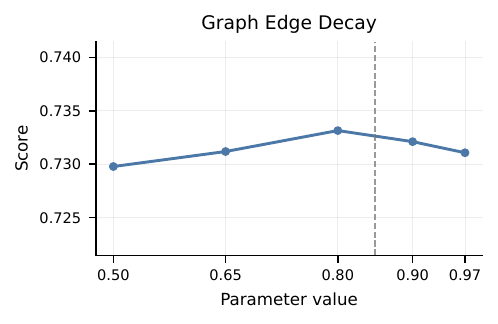}
\includegraphics[width=0.3\textwidth]{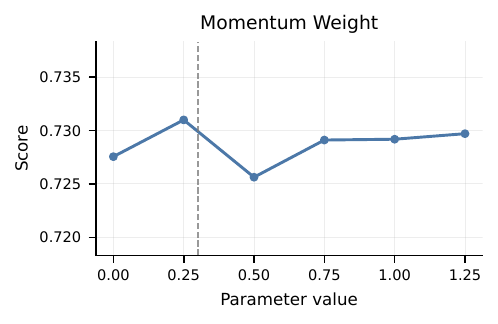}
\includegraphics[width=0.3\textwidth]{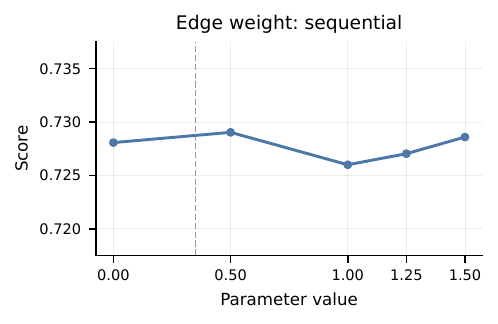}
\includegraphics[width=0.3\textwidth]{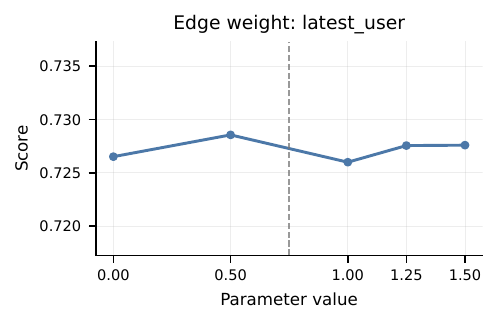}
\includegraphics[width=0.3\textwidth]{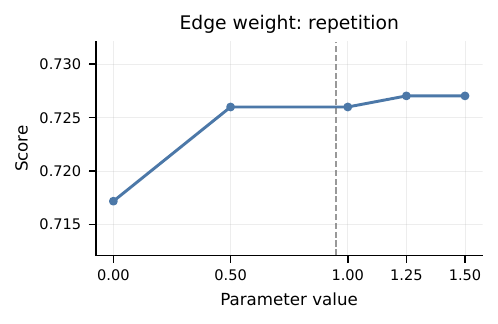}
\includegraphics[width=0.3\textwidth]{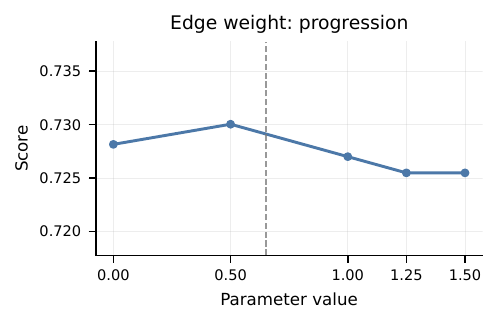}
\includegraphics[width=0.3\textwidth]{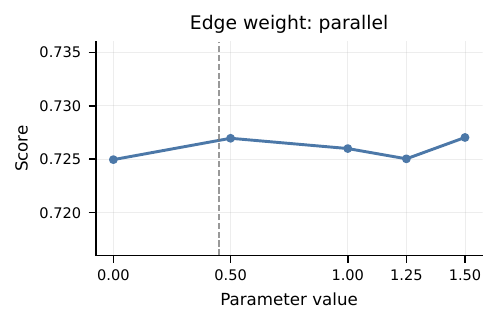}
\includegraphics[width=0.3\textwidth]{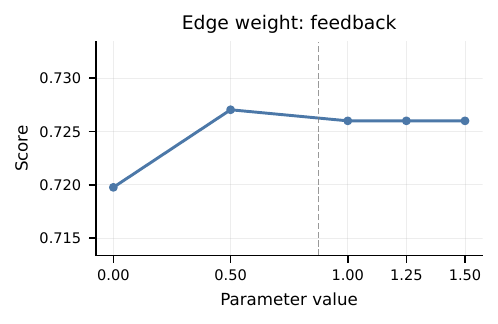}
\end{center}
\caption{Parameter sensitivity analysis of RUPA on GAIA with MiniMax-M2.7. Each subplot shows AUROC as a function of one hyperparameter while keeping the remaining settings fixed. The dashed gray line indicates the default values of hyperparameters, or the corresponding default edge weight used in the main experiments.}
\label{fig:parameter}
\end{figure}

\subsection{Detailed Experiment Settings}
\label{app:implementation}

All methods are evaluated under the same benchmark splits, prompts, and execution framework as described in the main paper. We use the Harbor framework for trajectory execution and verification, and fix the decoding temperature to 0.7 for all repeated-sampling based
methods. Unless otherwise stated, each repeated-sampling baseline is run with 3 samples per query, and the final uncertainty score is computed according to the original scoring rule of the corresponding method.

For baseline reproduction, we follow the official implementation whenever available. In particular, \textbf{Tracer} is reproduced using its released codebase. For \textbf{SAUP} and \textbf{Uprop}, no official implementation was available at the time of experimentation, so
we reimplemented both methods independently according to the descriptions in their papers and matched their reported scoring procedures as closely as possible. To ensure comparability, all baselines are evaluated on the same agent trajectories, model outputs, and task
instances as RUPA. For methods requiring step-level or trajectory-level aggregation, we preserve the original aggregation strategy specified by each baseline. Hyperparameters that are not explicitly defined by a baseline are set to the paper default when available;
otherwise, we use a validation-based choice on the training split without accessing test labels.

For RUPA, graph construction and uncertainty propagation use the outcome-blind calibration procedure described in the main text. All graph-related parameters, including relation weights, temporal decay, and history window size, are determined from unlabeled training
trajectories only, and is fixed across experiments of different model families and datasets. No test labels are used during parameter selection or calibration.

\subsection{Parameter Sensitivity Ablation Analysis}

In order to evaluate how the hyperparameter chosen in RUPA method afffect the final failure prediction performance, we conduct a parameter sensitive ablation analysis experiment on MiniMax-M2.7 model with gaia datasets, when one parameter ablation experiment is conducted, other hyperparameter is fixed as our main experiment setting. The result is shown in Fig ~\ref{fig:parameter}:

As shown by the resulting curves, for hyper-parameters like graph decay or momentum weight, RUPA is not overly sensitive to small perturbations around the default configuration, and its performance remains stable across a broad range of reasonable settings. Furthermore, the default hyperparameters consistently fall near a strong or near-optimal region for most parameters, suggesting that our edge-weight assignment strategy provides a sensible balance between different structural signals. These results indicate that the proposed parameterization is reasonable and that the edge-weight calibration method can assign meaningful importance to different relation types.

\end{document}